\def\year{2019}\relax
\documentclass[letterpaper]{article} 
\usepackage{aaai19}  
\usepackage{times}  
\usepackage{helvet}  
\usepackage{courier}  
\usepackage{url}            
\usepackage{booktabs}       
\usepackage{amsfonts}       
\usepackage{nicefrac}       
\usepackage{microtype}      

\usepackage{graphicx}
\usepackage{algpseudocode}
\usepackage{wrapfig}
\usepackage{subcaption}
\usepackage[textsize=small]{todonotes}
\usepackage{multirow}
\usepackage{amsmath}
\usepackage{verbatim}
\usepackage{threeparttable}
\begin{document}

\title{Logical Rule Induction and Theory Learning Using Neural Theorem Proving}

%

\author{
  Andres Campero \\ 
  MIT-IBM Watson AI Lab \\
  Brain and Cognitive Sciences, MIT \\
  campero@mit.edu\\
  \And
  Aldo Pareja \\
  MIT-IBM Watson AI Lab, \\
  IBM Research \\
  aldo.pareja@ibm.com\\  
  \And
  Tim Klinger \\
  MIT-IBM Watson AI Lab, \\
  IBM Research \\
  tklinger@us.ibm.com\\ 
  \AND
  Joshua B. Tenenbaum \\
  MIT-IBM Watson AI Lab \\
  Brain and Cognitive Sciences, MIT \\
  jbt@mit.edu\\
   \And
  Sebastian Riedel \\
  University College London \\
  s.riedel@cs.ucl.ac.uk\\ 
}
\nocopyright
\maketitle
\begin{abstract}

A hallmark of human cognition is the ability to continually acquire and distill observations of the world into meaningful, predictive theories. In this paper we present a new mechanism for logical theory acquisition which takes a set of observed facts and learns to extract from them a set of logical rules and a small set of core facts which together entail the observations.  Our approach is neuro-symbolic in the sense that the rule predicates and core facts are given dense vector representations.  The rules are applied to the core facts using a soft unification procedure to infer additional facts.  After $k$ steps of forward inference, the consequences are compared to the initial observations and the rules and core facts are then encouraged towards representations that more faithfully generate the observations through inference. Our approach is based on a novel neural forward-chaining differentiable rule induction network.  The rules are interpretable and learned compositionally from their predicates, which may be invented. We demonstrate the efficacy of our approach on a variety of ILP rule induction and domain theory learning datasets.
\end{abstract}

\section{Introduction}
\begin{figure}
  \centering
  \includegraphics[width=0.48\textwidth]{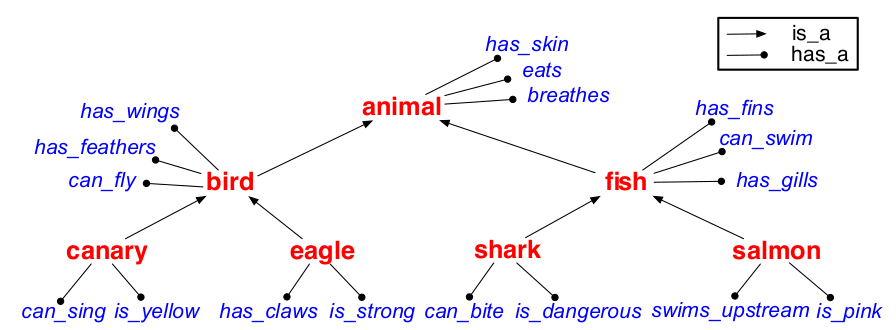}
  \caption{Animal Taxonomy. 
  Constants are in red and blue, relations are indicated with lines and arrows.}
  	\label{fig:taxonomy_small}
\end{figure}

Humans are continually acquiring, representing, and reasoning with new facts about the world. To make sense of the vast quantity of information with which we are presented, we must compress, structure and generalize from what we experience. This allows us to quickly understand new concepts and make useful predictions about them.  For example, we might represent our knowledge of animals in a taxonomic hierarchy like that shown in Figure \ref{fig:taxonomy_small}.  Using such a hierarchy coupled with an inheritance rule that specifies that the attributes of higher nodes are shared by lower ones, we can achieve exponential compression over a representation which just lists the facts.  Even more exciting, it allows us to infer a whole range of new facts about an individual simply by observing where it fits in the hierarchy.  For example, observing that a Harpy Eagle is a type of Eagle allows us to immediately deduce that a Harpy Eagle can fly and breathe.\footnote{There are some kinds of reasoning which are not easy to do with a taxonomy (for example, handling the exception that penguins are birds but don't fly) but our proposal is not limited to taxonomic representations.} But how can such representations be learned from raw observations?  This has been a key problem in semantic knowledge acquisition going back to at least to the 1960's in the work of \citeauthor{collins69} \shortcite{collins69}, with symbolic, Bayesian, and neural approaches proposed \cite{rogersm04,hinton1986learning,katz2008}. In our view (and following \citeauthor{katz2008} \citeyear{katz2008}) there are three questions to be addressed in the development of a solution: (1) how can we induce logical rules from the observations? (2) how can we learn a small set of core facts (the taxonomy in the example) from which we can infer the observations (and more), and (3) how can this be done without explicit supervision on the structure of the rules?

In this paper we propose a model which can be used for both Inductive Logic Programming (ILP) and theory acquisition/compression. The network is neuro-symbolic in the sense that it represents predicates for both rules and facts using dense vectors which can be trained using gradient descent towards representations of known predicates (including a fixed set of anonymous invented predicates).  The network implements forward chaining and soft unification. \footnote{Soft unification relaxes the requirement that two predicate symbols must be identical for the rule to be applicable instead favoring a measure of the degree of similarity.}

For ILP problems, the network is given a set of ``proto-rules" (rules with randomly initialized predicate parameters) and applies them using forward chaining to the background facts to produce consequent facts.  After $K$ steps of forward chaining ($K$ is a hyper-parameter) the consequent facts are compared to the labeled target facts and the rule predicate parameters are trained towards representations of the predicates which yield all the true target facts and none of the false ones. Representations can be learned either for the known predicates used in the facts or for auxiliary invented predicates. 

In theory acquisition/compression, the network is given a set of fact observations and asked to learn a logical theory -- a set of core facts and a set of rules -- which together entail the observations. The ability to learn facts is an aspect that has not been emphasized in many ILP approaches but is present in the Bayesian literature. For example, when observing that salmon can swim, have fins and have gills, the model can learn the core fact that salmon are fish even though that is not deducible directly. By encouraging sparsity in the set of core learned facts with a penalty term in the loss, the model can be trained to try to minimize the size of the theory it learns.

The remainder of the paper is structured as follows. In the next section we provide background material and related work.  Then we present our model and training procedure.  In Section 4 we present experiments which investigate different capabilities of the model and demonstrate the efficacy of our approach on a variety of ILP rule induction and domain theory learning datasets. We conclude with a discussion of limitations and future directions.

\vspace{.3cm}
\section{Background and Related Work}
%
%

There is a rich literature on neuro-symbolic induction to which our approach is related on two main lines: inductive logic programming (ILP) and semantic cognition.
ILP systems try to learn a set of logical rules which can be used to deduce facts of interest while semantic cognition is broadly concerned with how human beings acquire, represent, and integrate knowledge.

\subsection{Inductive Logic Programming}
In ILP, the goal is to learn (induce) logical rules which can be chained to successfully answer queries about a target relation, given positive and negative examples of that relation and some background facts. Logical rules are of the form:
\begin{equation}
h \leftarrow b_1, b_2, \ldots, b_k
\end{equation}
where $h$ is an atom called the \emph{head} of the rule and $b_1, b_2, \ldots, b_k$ are atoms which constitute the \emph{body}.  An \emph{atom} is a positive or negative literal.  A literal is a predicate applied to terms which may, in our case, be either variables or constants.  For example $grandfather(X,Y)$ is a (positive) atom whose predicate is $grandfather(\cdot, \cdot)$ and whose arguments (two in this case) are variables $X$ and $Y$.  When the arguments of the atom are all constants (e.g. $parent(Tom, Bill)$ for constants $Tom$ and $Bill$) we call it a \emph{ground atom} which we also refer to as a \emph{fact} when it is given as true or its truth value is inferred from rules. Intuitively, the head of the rule is true if each of the $b_i$ in the body are. For example a rule might be:
$$grandfather(X, Y) \leftarrow father(X, Z), parent(Z, Y)$$

Which is read: \emph{$X$ is the grandfather of $Y$ if $X$ is the father of $Z$ AND $Z$ is the parent of $Y$}.  The head atom holds for any $X$ and $Y$ as long there is some individual $Z$ for which $father(X,Z)$ and $parent(Z,Y)$ are both true facts.  

Given background facts such as $father(Bill, Mary)$ and $parent(Mary, Liz)$, logical rules can be chained together to prove a goal fact like $grandfather(Bill, Liz)$.  This is an example of a \emph{forward chaining} deduction because it starts from a set of facts and unifies (matches) them with the body of a rule to derive the consequences.  It is also possible to do \emph{backward chaining} in which we start with a goal and work backwards by unifying it with the head of a rule, recursively trying to prove the body.

Symbolic ILP systems have a rich history dating back decades.  A common approach to inducing rules is called \emph{learning from entailment} \cite{DeRaedt:1997:LSC:270588.270596}, in which hypothesized rules are combined with background facts and trained to entail the positive and none of the negative concept examples. The FOIL algorithm \cite{reason:QuiCam95a} is an example of this approach. Our approach is also of this sort though we use neural networks to learn the rules. The classic ILP setting has been continually updated to handle richer knowledge.  In   \cite{de2008probabilistic} for example they provide several formulations of probabilistic ILP.  We do not consider probabilistic interpretations in our approach, though that is an interesting avenue for future research.  

A related branch of research called Abductive Logic Programming attempts to learn consistent explanatory facts as well as rules \cite{kakas1992abductive}.  For example, it might allow us to induce the fact that eagles are birds from the facts that eagles have wings and feathers together with the inheritance rule. Our approach for theory acquisition may be considered an example of this line of work.

\subsection{Neuro-Symbolic Integration}
Symbolic ILP systems do very well at generalizing from just a few examples.  This is because they learn universal rules.  They are, however, susceptible to noisy inputs and even a single bad fact can cause them to fail.  On the other hand, neural systems generally are very robust to noisy input but are sample inefficient and prone to over-fitting on small amounts of data.  Neuro-symbolic systems aim for the best of both worlds.  They can be made robust to noisy inputs while still retaining some of the strong generalization properties typically associated with symbolic systems. 

There is a long history of research in neural-symbolic systems from which we choose just a small set to present here.  For a recent survey see \cite{DBLP:journals/corr/abs-1711-03902}. 

In \cite{serafini2016logic} they introduce Logic Tensor Networks (LTN) and Real Logic whose semantics grounds the terms, atoms and clauses of the language as continuous functions.  They demonstrate that the logic can be implemented using neural networks for the groundings of the symbols and apply it to solving a database completion problem.  A follow-up paper applies LTN's to semantic image segmentation \cite{donadello2017logic}. Neither work considers the problems of rule induction or theory acquisition. 

A recent work \cite{manhaeve2018deepproblog} starts from a probabilistic logic programming language (problog) and extends it to handle neural predicates which compute probabilities.  Like Prolog and ProbLog, DeepProbLog is a backward chaining approach.  It leverages the automatic differentiation system of ProbLog to incorporate neural predicates and trains with gradient descent. Like Logic Tensor Networks, ProbLog can train neural network implementations of relations.  ProbLog does not do rule induction.    

In Neural LP \cite{yang2017differentiable}, the system can learn chaining-type rules. It uses a neural controller built on top of TensorLog \cite{DBLP:journals/corr/Cohen16b} and is trained to learn rules to compute a ranked list of entities which satisfy a partially specified query. It differs from our approach in several respects. It requires a partially specified query. It represents predicates as TensorLog operators (matrices) whereas we represent them as parameter embeddings which can be associated with constants and a valuation to represent an atom. And it is not obvious how it could be applied to learn fact representations. 

\cite{sourek2015lifted} uses templates to create grounded networks that depend on the example. \cite{tran2016deep} studies the incorporation and extraction of knowledge into deep networks. \cite{kazemi2017relnn} focuses on predicting the properties of objects.

Our approach is most directly related to two recent neural ILP approaches. In \cite{DBLP:journals/corr/abs-1711-04574} inference is done through the forward chained application of a set of logical rules. During learning, a set of all the possible candidate rules is generated according to a provided template. Parameters are weights associated with pairs of candidate rules. These weights are normalized to lie in $[0, 1]$ and interpreted as probabilities associated to the rule pairs as possible definitions of the concept. When there are a large number of rules, this method may suffer scalability issues. In addition it requires a representation of the truth values of all possible facts and non-facts.  By contrast,  \cite{DBLP:journals/corr/RocktaschelR17}  construct a function representing a backward-chained proof of the goal and require only a representation of the true facts.  A more conceptual distinction arises in their parameterizations.  In \cite{DBLP:journals/corr/abs-1711-04574} the parameters are weights on rule pairs.  In \cite{DBLP:journals/corr/RocktaschelR17} they start with a set of parameterized rules which, as in our approach, acquire their meaning as the predicate embeddings of their head and body atoms are trained through unification with the predicates of the facts.  It is not obvious how these approaches could be applied to theory acquisition requiring fact induction.

In our approach we follow \cite{DBLP:journals/corr/RocktaschelR17} in parameterizing with embeddings but use forward rather than backward chaining so that we don't have to represent a proof tree explicitly. Unlike \cite{DBLP:journals/corr/abs-1711-04574}, we don't have to generate all the candidate rules. Instead, learning is at the level of individual atoms.

\subsection{Semantic Cognition}
Semantic cognition concerns the acquisition and integration of knowledge. Previous work has modeled semantic cognition as a kind of logical dimensionality reduction \cite{katz2008,ullman2012theory} which uses probabilistic generative models that can simultaneously learn logical rules and a set of core relations that form a theory underlying the observed data. Like ILP approaches, these models can make deductive inferences through the application of logical rules. But unlike traditional ILP algorithms, these Bayesian models are also able to induce facts. 

This ability to apply both inductive and deductive reasoning at the level of both facts and rules provides humans with a rich space of techniques with which to tame the complexity of everyday experience. These approaches illuminated a promising direction but were severely challenged by scalability issues.  With the success of neural techniques we believe it is useful to revisit these ideas.

\begin{figure*}[t]
  \centering
  \includegraphics[scale=0.33]{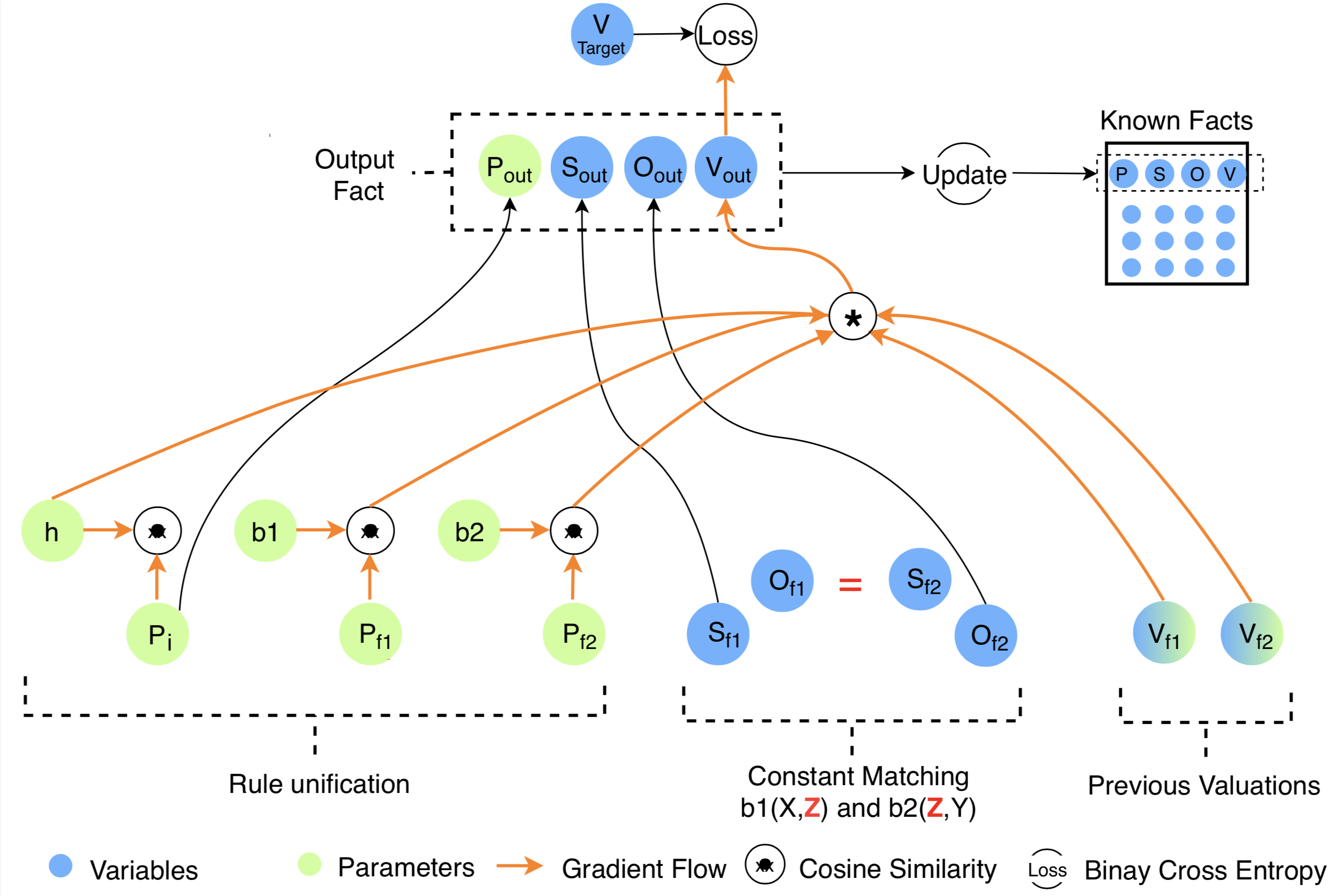}
  \caption{Overview of the model, a step of forward chaining. Parameters are represented in green  and constitute the trainable embeddings, orange arrows indicate paths on which gradients flow (in the opposite direction).}
  	\label{fig:model}
\end{figure*}


\section{The Model}

As described above, our focus is on two tasks: rule induction in an ILP setting, and theory learning (learning both core facts and rules).  In this section we describe a model which can be configured to perform either task and is trained using stochastic gradient descent.  Figure \ref{fig:model} illustrates the architecture common to both tasks.  A rule is shown on the bottom left with a head predicate $h$ and two body predicates $b1$ and $b2$.  The model represents the rule as a triple $((\mathbf{\theta}_h, v_1, v_2), (\mathbf{\theta}_{b1}, v_3, v_4), (\mathbf{\theta}_{b2}, v_5, v_6))$. $\mathbf{\theta_{h}}$, $\mathbf{\theta_{b1}}$, and $\mathbf{\theta_{b2}}$ are parameterized embeddings  in $\mathbf{R}^d$ corresponding to predicate symbols $h, b_1, b_2$; $v_i$ denote variables $X, Y, Z$ which are the subjects and objects that form the arguments of the atoms. 

Facts are represented as quadruples $(\mathbf{\theta}_p, s, o, v)$, where $\mathbf{\theta}_p$ is a parameter vector shared by all the facts associated with predicate $p$; $s$ is the subject constant; $o$ is the object constant; and $v \in [0, 1]$ is a valuation representing the model's degree of belief in the truth of the atom $p(s, o)$.  Constants are internally represented as integers and may be mapped to symbols for interpretation. The set of all current known facts (either initially given or inferred) is shown in the figure at the upper right.  We maintain a set of all predicates $P$ and their associated embeddings which may include auxiliary predicates that can be used for predicate invention in the learning of rules and concepts. All predicates are randomly initialized.



To apply the rule to a pair of facts $f_1, f_2$, where $f_i = (\mathbf{\theta_{p_i}}, s_{f_i}, o_{f_i}, v_{f_i})$ we first check for constant matching between the structure of the rules and the constants of the facts. This is just checking that the variable arguments of the rule body can be assigned to the corresponding constants in the facts.  If they cannot, then then network construction stops.  For example, as in the figure, the rule $grandfather(X, Y) \leftarrow father(X, Z), parent(Z, Y)$ can be applied successfully to the pair of facts $father(Bill, Tom), parent(Tom, Mary)$.  However, if applied to $father(Bill, Tom), parent(Anne, Mary)$, it would not match and would cause network construction to stop for this rule and fact pair.  After a successful constant matching, a set of candidate output facts is generated, one for each predicate $p \in P$. The arguments of the fact are determined by the bindings of the rule body predicates to the input facts. The figure illustrates this with arrows flowing from the constants of the input facts to $S_{out}$ and $O_{out}$ forming a consequent fact at the top.  The creation of separate facts for each predicate in $P$ is required because of our ignorance of the correct predicate for the head of the rule. 

The valuation for each candidate output fact is determined through soft unification by (differentiably) combining the values of the input facts with measures of the degree of similarity between each input fact predicate and the corresponding rule body embedding, as well as with the similarity between the rule head embedding and the candidate fact predicate in $P$.  The values of the input facts are multiplied to implement a soft form of AND \footnote{There are many other choices which are more theoretically well-grounded, such as the t-norm, but we found that simple multiplication works the best in practice for this application.  Note that because we restrict ourselves to at most two atoms in the body of a rule, there is no issue with underflow here.}. Specifically, we compute the value $v_{out}$ of an output fact $(\mathbf{\theta}_p, s_{out}, o_{out}, v_{out})$ resulting from the unification between a rule with head and body predicates $\mathbf{\theta_h}$, $\mathbf{\theta_{b_1}}, \mathbf{\theta_{b_2}}$ and facts $f_1$ and $f_2$ with values and corresponding predicates $v_{f_1}$, $v_{f_2}$, $\mathbf{\theta_{f_1}}$ and $\mathbf{\theta_{f_2}}$ as:
\begin{equation}
v_{out} = \cos(\mathbf{\theta}_h, \mathbf{\theta}_{p}) \cdot \cos(\mathbf{\theta_{b_1}}, \mathbf{\theta_{f_1}}) \cdot \cos(\mathbf{\theta_{b_2}}, \mathbf{\theta_{f_2}}) \cdot v_{f_1} \cdot v_{f_2}
\end{equation}


If the predicate and arguments of a consequent fact matches one in the set of known facts, its value is updated with the max between its previous and newly inferred values (implementing an OR); if it is not, the new fact is appended to the set of known facts. In this way the valuation is dynamically extended at each step of inference.

We have described how the network is constructed to perform one step of inference with a single rule.  To construct the entire network, we start with an initial set of facts and perform this inference procedure for each rule and for each pair of facts.  If a pair of facts fails to unify with the rule then that branch of the construction terminates.  The number of steps of inference $K$ is a hyper-parameter as is the number of auxiliary predicates included in the predicate set $P$.

To train the network we use a loss function that depends on the task (we describe the setup for each task below).  The loss gradients are back-propagated to update the predicate embeddings for the rules and for the facts (the predicates of the facts can also be fixed, i.e as one-hot vectors). The orange arrows of Figure \ref{fig:model} indicate the paths on which gradients flow (in the opposite direction).   The rule and fact predicate embeddings are the parameters of the network, shown in green.

When a set of background facts is given, as in the case of the ILP tasks, we initialize the current valuation for the known facts  to 1.0 and train the rules and predicates using a binary-cross-entropy loss to produce the correct values for the positive and negative target facts.

For theory learning, the aim is to learn a small theory which can recover the observations and generalize using the logical rules. Thus, we additionally learn a set of initial core facts that underlie the structure of the observations.  In order to do that, unlike in the ILP setting, we additionally parameterize the valuations for all the facts and initialize them to 0.5 reflecting our initial ignorance of their truth.  We  train them towards values which allow the model to faithfully recover the observations.  The model may produce additional facts not in the observations but the loss penalizes only the implied facts for whose predicates and arguments match those of an observed fact but for whose values differ.  A regularization term controlled by $\lambda$ penalizes the squared sum of the initial core valuations, encouraging compression.  Using the notation $i \sim f$ for facts $f$ and $i$ to indicate that their predicates and arguments match exactly, and $v(f)$ to denote the value of a fact $f$, the loss can be written:
\begin{equation}
\sum_{i \in I , f \in F, i \sim f} BCE(v(f), v(i)) + \lambda\sum_{i \in I} v(i)
\end{equation}

\section{Experiments}

A wide range of previous work has focused on different aspects of logical induction. Here we test the capabilities of our algorithm in three different settings. First we test the capability of the algorithm to perform logical rule induction in the set of tasks covered by \cite{DBLP:journals/corr/abs-1711-04574}. 
Second, we test our algorithm in a bigger dataset and compare with \cite{DBLP:journals/corr/RocktaschelR17}.
Finally, we test our model in the context of learning domain theories following \cite{katz2008,ullman2012theory}, the algorithm has to simultaneously learn the atoms of the logical rules, the representations of the facts of the predicates, and a set of core facts. To do that the algorithm learns to do deductive and inductive inferences of the facts.

\subsection{Rule Learning ILP Tasks}
We tested the model in the ILP problems from  \cite{DBLP:journals/corr/abs-1711-04574}\footnote{We only skip the Husband and Uncle tasks which require the datasets from \cite{wang2015soft}.}. The task of an ILP problem is to learn a target relation given a set of background knowledge facts $\mathcal{B}$ and a set of positive $\mathcal{P}$ and negative $\mathcal{N}$ examples of a target relation.

As an example, consider the task of learning the predicate \textit{even(X)}. The background knowledge is defined using the \textit{zero(X)} and \textit{succ(X,Y)} predicates.
$$\mathcal{B}=\{\textit{zero}(0),\textit{succ}(0,1),\textit{succ}(1,2),... ,\textit{succ}(9,10)\}$$

The target positive and negative predicate extensions are:
$$\mathcal{P} = \{\textit{target}(0), \textit{target}(2),...,\textit{target}(10) \}$$
$$\mathcal{N} = \{\textit{target}(1),\textit{target}(9)\}$$

An example solution found by the algorithm is:
$$target(X)\leftarrow zero(X)$$
$$target(X)\leftarrow target(Y), auxpred(Y,X)$$
$$auxpred(X,Y)\leftarrow succ(X,Z), succ(Z,Y)$$
Where \textit{auxpred} acquires the meaning of \textit{succ2} which is true of $X,Y$ whenever $X+2=Y$.

Table \ref{evans_res} gives a performance comparison to \cite{DBLP:journals/corr/abs-1711-04574}. Since universal logical rules are perfectly generalizable, and to facilitate comparison, we use the same evaluation metric: the percentage of runs with different random weight initializations that successfully learn rules to solve the task with less than $1e-4$ mean squared error. To avoid local minima, we explored adding a decaying normal noise to the embeddings. This had a small positive effect in some of the tasks, reported results include the effect of the noise. Details of the problems are given in the Appendix.

\begin{table}[!ht]
 \caption{ILP percentage of successful runs. $|I|$ is the number of intentional predicates.}
 \label{evans_res}
 \centering
 \begin{tabular}{lllll}
   \toprule
	
   \multirow{2}{*}{Task}      & \multirow{2}{*}{$|I|$}  & \multirow{2}{*}{Recursive} &\multirow{2}{*}{$\partial ILP$}  &\multirow{2}{*}{Ours} \\
   
    \\

   \midrule
   Predecessor & 1 & No & 100 & 100\\
   Even-Odd & 2 & Yes & 100 & 100\\
   Even-succ2 & 2 & Yes & 48.5 & 100\\
   Less than & 1 & Yes & 100 & 100\\
   Fizz & 3 & Yes & 10 & 10\\
   Buzz & 2 & Yes & 35 & 70\\
   Member & 1 & Yes & 100 & 100\\
   Length & 2 & Yes & 92.5 & 100\\
   Son & 2 & No & 100 & 100\\
   Grandparent & 2 & No & 96.5 & 100\\
   Relatedeness & 1 & No & 100 & 100\\
   Father & 1 & No & 100 & 100\\
   Undirected Edge & 1 & No & 100 & 100\\
   Adjacent to Red & 2 & No & 50.5 & 100\\
   Two Children & 2 & No & 95 & 0\\
   Graph Colouring & 2 & Yes & 94.5 & 0\\
   Connectedness & 1 & Yes & 100 & 100\\
   Cyclic & 2 & Yes & 100 & 100\\

   \bottomrule
 \end{tabular}
\end{table}

We see that our algorithm performs equally or better in most of the tasks.  In contrast to theirs, search is not made at the level of rules but at the more compositional level of atoms. In fact, when the embeddings of the dictionary of predicates are fixed as one-hot vectors, our procedure is very similar to theirs: the parameters that form the predicates of the rules can be treated as weights that select the correct predicate. Thus, like in their model, training consists on selecting the weights that make the embeddings look like the right one-hot vectors and the procedure becomes a symbol search, except at the level of the atoms instead of the rules. The more general case where the embeddings are trained and dense, opens the interesting direction to be explored of studying the learned vector embedding semantic space, as has been done for standard NLP tasks \cite{mikolov2013efficient}, which can potentially allow for similarity and analogical reasoning. In our table, we report the result with the trainable embeddings.

It is worth noting the failure of our model in the Graph Colouring and in the Two Children tasks. A quick exploration suggests that the global optima has a very sharp neighborhood while the local minima are attractors in most of the space. This is reminiscent of the Terpret problem \cite{terpretproblem,gaunt2016terpret} and the local minima can only be avoided when the random initialization is very close to the correct rules.

\begin{table*}[!ht]
  \hspace*{1.5cm}
 \begin{threeparttable}
 \caption{Performance on the COUNTRIES dataset}
 \label{sebastians}
 \centering
 \begin{tabular}{l|c c c|l}
   \toprule
	
   \multirow{2}{*}{Task}      & \multicolumn{3}{c|}{Model AUC-PR}  & \multirow{2}{*}{Rule examples and confidences}  \\
      \cmidrule(r){2-4}

& NTP & NTP-$\lambda$ & Ours & \\

   \midrule
   \multirow{2}{*}{S1} & \multirow{2}{*}{$90.83 \pm 15.4$ } & \multirow{2}{*}{$100.00 \pm 0.0$} & \multirow{2}{*}{$91.15 \pm 15.4$} & \multirow{2}{*}{0.85 loc(X,Y) $\gets$ loc(X,Z), loc(Z,Y)}  \\
   &&&&\\
   \multirow{2}{*}{S2} & \multirow{2}{*}{$87.40 \pm 11.7$ } & \multirow{2}{*}{$94.04 \pm 0.4$ } & \multirow{2}{*}{$86.87 \pm 3.2$}  & \multirow{2}{*}{0.57 loc(X,Y) $\gets$ neighbor(X,Z), loc(Z,Y)}  \\
   &&&&\\
   \multirow{2}{*}{S3} & \multirow{2}{*}{$56.68 \pm 17.6$  } & \multirow{2}{*}{$77.26 \pm 17.0$ } & \multirow{2}{*}{$63.08 \pm 28.2$ }  & 0.59 loc(X,Y)  \\
  &&&&$\gets$ neighbor(X,Z), loc(Z,W), loc(W,X) \\
   \bottomrule
 \end{tabular}
 \end{threeparttable}
\end{table*}

Notice that since our approach is differentiable, it is not prone to some of the problems of symbolic ILP, and like in \citeauthor{DBLP:journals/corr/abs-1711-04574}, it can handle ambiguity and noise.

\subsubsection{Countries}
We are not focused specifically on knowledge base completion but use the COUNTRIES dataset \cite{bouchard2015approximate} to evaluate the scalability of our algorithm, comparing to other neural logical approaches. The dataset contains 272 constants, 2 predicates and 1158 true facts and is designed to explicitly test the logical rule induction and reasoning capabilities of link prediction models. 

We compare on the 3 tasks described in \cite{DBLP:journals/corr/RocktaschelR17}, requiring reasoning steps of increasing length and difficulty (S1,S2,S3 in table \ref{sebastians}). We report the Area Under the Precision-Recall-curve (AUC-PR) where results are comparable to the previous NTP approach. For completion, we also report $NTP-\lambda$ from the same paper which uses an additional neural link network as an auxiliary loss. Like them, we also show some example rules and a confidence score by taking the minimum similarity between the atoms of the rule and their decoded predicates.

To perform the forward chaining during training, at each epoch we randomly sample from a section of the knowledge graph both the targets and a set of facts to form the background knowledge. Like the related work, our model can also suffer from scalability issues, as in forward chaining the size of the facts grows exponentially with the number of steps. Restricting the number of considered facts through sampling was sufficient for the task at consideration but this could show problems when scaling to much bigger datasets, we discuss some future directions in the conclusion.

\begin{table*}[t]
  \hspace*{1.7cm}
 \begin{threeparttable}
 \caption{Theory Learning Results. Succ is the percentage of successful initializations; Acc stands for the accuracy of the recovered facts; Const is the number of constants. }
 \label{theory}
 \centering
 \begin{tabular}{ll|ccc|ccc}
   \toprule
   &&\multicolumn{3}{c|}{Taxonomy}  &\multicolumn{3}{|c}{Family}                 \\
	\midrule
    &&\# Preds & \# Const & \# Facts &\# Preds & \# Const & \# Facts\\
    \cmidrule(r){3-8}
    \multicolumn{2}{l|}{\multirow{2}{*}{Observed Data} }&\multirow{2}{*}{4}&\multirow{2}{*}{36}&\multirow{2}{*}{145}&\multirow{2}{*}{6}&\multirow{2}{*}{10}&\multirow{2}{*}{30}\\
   &&&&&& \\
    \multicolumn{2}{l|}{Target Theory}&4&36&40&4&10&28\\
    \midrule
    &&\% Succ & \% Acc & \# Induced Facts &\% Succ & \% Acc & \# Induced Facts\\
    \cmidrule(r){3-8}
    \multicolumn{2}{l|}{\multirow{2}{*}{Algorithm}}&\multirow{2}{*}{70}&\multirow{2}{*}{99}&\multirow{2}{*}{69}&\multirow{2}{*}{100}&\multirow{2}{*}{96}&\multirow{2}{*}{30.8}\\
    &&&&&&\\
   \bottomrule
 \end{tabular}
 \end{threeparttable}
\end{table*}

\subsection{Learning Theories}
We test the capability of our network to compress a set of observations in the form of a theory by learning a set of core facts in addition to the logical rules. We take the two domains considered by \cite{katz2008}: Taxonomy and Kinship.

\subsubsection{Taxonomy}\label{taxonomy}
A taxonomy is a set of observations structured into a tree where downstream nodes inherit the properties from the nodes above them in the tree. As shown in Figure 1, the tree constitutes a theory formed by two logical rules $IS(X,Y) \leftarrow IS(X,Z), IS(Z,Y)$; $HAS(Z,Y) \leftarrow IS(X,Z), HAS(Z,Y)$ capturing inheritance  and by a set of core facts represented with the edges. All observed facts can be recovered by iterative application of the rules (if salmon are fish, and fish have gills, then salmon have gills).

From a range of different possible taxonomies we report performance on the bigger original one from \cite{rogersm04}. This data contains 145 facts composed of 4 predicates and 36 constants. The facts can be compressed into a tree structured theory as shown in the  Appendix that contains only 40 core facts.

As shown in Table \ref{theory}, the algorithm is able to learn the theory in 70\% of the runs, achieving 99\% accuracy and compressing close to the optimal level (average of 69 compared to the optimal of 40).

\subsubsection{Kinship} \label{kinshipSec}
We also evaluated performance on the difficult kinship theory, which contains 10 constants and 6 observed predicates \textit{mother}, \textit{father}, \textit{daughter}, \textit{wife}, \textit{husband} (see figure in the Appendix). In this case the compression of the theory consists of a set of 4 auxiliary core predicates with 28 facts. The algorithm has to learn the concepts \textit{female}, \textit{male}, \textit{spouse}, \textit{child} which acquire their meaning through their extensions and the 6 logical rules that generate the observations:
$$ mother(X,Y)\leftarrow female(X), child(Y,X)$$
$$ father(X,Y)\leftarrow male(X), child(Y,X)$$
$$ daughter(X,Y)\leftarrow female(X), child(X,Y)$$
$$ son(X,Y)\leftarrow male(X), child(X,Y)$$
$$ wife(X,Y)\leftarrow female(X), child(X,Y)$$
$$ husband(X,Y)\leftarrow male(X), child(X,Y)$$

Table 3 shows the statistics for the observed and target compressed data. The algorithm's performance is again quantified as the percentage of initializations where the rules are successfully learned, the accuracy of the recovered data and the number of learned core facts. The algorithm is able to perform this compression and learns a set of new predicates that conform the rules, recovering 96\% of the data correctly (the algorithm sometimes deduces that some facts are true when they aren't because it can induce incorrect core facts).

\section{Conclusions and Future Work}
We have presented a forward chaining inference network which is parameterized in the embeddings that form its rules and facts.  Learning in this model means simultaneously learning the right sub-symbolic representations, and the right resulting symbolic conceptual relations implied through the logical rules; together constituting a Dual-Factor definition of the concepts \cite{carey2009origin}.

 We articulated a set of desiderata for models which learn logical theories from observations which include: compression, rule induction, and the ability to learn without direct supervision. We showed how our inference procedure satisfies these three conditions in two settings. Our model was able to learn a significant compression for the taxonomy and kinship datasets proposed by \cite{rogersm04,katz2008}, learning interpretable representations not just for the parametrized rules but also for facts -- a feature lacking in many traditional ILP solutions -- using only the supplied observations as supervision. These are encouraging results for theory acquisition and point to the viability of this approach. As demonstrated on the ILP datasets from \cite{bouchard2015approximate,cropperlearning} and those from \cite{DBLP:journals/corr/RocktaschelR17}, the method also provides an interesting alternative in the ILP setting.


\subsubsection{Limitations}
As in previous work, our model is provided with rules that conform to templates, ideally this should not be necessary. The network needs to consider the set of all possible facts when doing core fact induction (which is not necessary for the rule induction problems), this is not be scalable in practice. Forward chaining grows exponentially in the number of facts considered at each step, this can also present a problem when scaling to bigger datasets. From a cognitive science perspective, the model is still more limited than its Bayesian symbolic counterparts, specifically, while those models provide graded measures of confidence in their inferences, our neural logical reasoner does not currently provide meaningful consistent estimates of uncertainty.


\subsubsection{Future Directions} 
One straightforward attempt of learning the structural information of the rules provided by the templates (arity and variable order) would be to encode it by adding dimensions to the embeddings and have the algorithm interpret them by using independent unifications in the desired way. This would constitute a slightly more complicated learning task but would maintain the same structure and mechanism that could be trained through gradient descent. It would also be interesting to explore richer sampling procedures and the integration of forward with backward chaining, this could perhaps yield regimes more similar to those of humans and could help scale to larger datasets. We would also like to investigate ways of providing better estimates of uncertainty -- from a full neural probabilistic formulation, to a heuristic metric based on the number of initializations and on the unification scores. 



\bibliography{aaai}
\bibliographystyle{aaai}

\newpage
\clearpage

\section{Appendix}

\subsection{ILP Tasks}

Description of tasks and the proto-rule templates used during training. More details are in \cite{DBLP:journals/corr/abs-1711-04574}
\subsubsection{Predecessor}
In this task we aim to learn the predecessor relation, $predecessor(X,Y) \leftarrow succ(Y,X)$, from basic aritmetic facts $\{zer(0),succ(0,1),suc(1,2), ...\}$. We use the following template:
\begin{align*}
F(X,Y) &\leftarrow F(Y,X) & (9)
\end{align*}

\subsubsection{Even-Odd}
In this task we aim to learn the $even$ predicate. Here the background knowledge is the same as in Predecessor (above). We must include an extra auxiliary predicate that learns to encode the relation $odd$. We use the following templates:
\begin{align*}
F(X) &\leftarrow F(X) & (1)\\
F(X) &\leftarrow F(Z), F(Z,X) &(2)\\
F(X) &\leftarrow F(Z), F(Z,X) &(2)
\end{align*}

\subsubsection{Even-succ2}
Described in Experiments.
\begin{align*}
F(X) &\leftarrow F(X) & (1)\\
F(X) &\leftarrow F(Z), F(Z,X) &(2)\\
F(X,Y) &\leftarrow F(X,Z), F(Z,Y) & (3)
\end{align*}

\subsubsection{Less Than}
Here we aim to learn the $lessThan$ relation. Background knowledge is the same as in the tasks above. A possible solution would be:\\
\\
$lessThan(X,Y) \leftarrow succ(X,Y)$ \\
$lessThan(X,Y) \leftarrow lessThan(X,Z), lessThan(Z,Y)$\\
\\
We used these templates:
\begin{align*}
F(X,Y) &\leftarrow F(X,Y) & (5)\\
F(X,Y) &\leftarrow F(X,Z), F(Z,Y) &(3)
\end{align*}

\subsubsection{Fizz}
As in the children game Fizz-Buzz, numbers that are divisible by three should be classified as Fizz. For more details refer to \cite{DBLP:journals/corr/abs-1711-04574}
These are the template protorules used during training:
\begin{align*}
F(X) &\leftarrow F(X) & (1)\\
F(X) &\leftarrow F(Z), F(Z,X) &(2)\\
F(X,Y) &\leftarrow F(X,Z), F(Z,Y) & (3)\\
F(X,Y) &\leftarrow F(X,Z), F(Z,Y) & (3)
\end{align*}

\subsubsection{Buzz}
Following the same logic as in Fizz, we used the following templates:
\begin{align*}
F(X) &\leftarrow F(X) & (1)\\
F(X) &\leftarrow F(Z), F(Z,X) &(2)\\
F(X,Y) &\leftarrow F(X,Z), F(Z,Y) & (3)
\end{align*}

\subsubsection{Member}
Here we aim to learn $member(X,Y)$, which is true if $X$ is an element of list $Y$. The background knowledge encodes values on a list by using two predicates: $cons(X,Y)$ which is true if node $Y$ is after list $X$ (lists are terminated with the null node 0); and $value(X,Y)$ which is true if the value of node $X$ is $Y$. One possible solution is:\\
\\
$member(X,Y) \leftarrow value(Y,X)$\\
$member(X,Y) \leftarrow cons(Y,Z), member(X,Z)$\\
\\
We used these templates:
\begin{align*}
F(X,Y) &\leftarrow F(Y,X) & (9)\\
F(X,Y) &\leftarrow F(Y,Z), F(X,Z) &(10)
\end{align*}

\subsubsection{Length}
The $length(X,Y)$ relation is true if the length of list $X$ is $Y$. We represent lists in the same way as in the $Member$ task. We required at least one extra intentional predicate $pred1$. One possible solution would be:\\
\\
$Length(X,X)\leftarrow zero(X)$\\
$Length(X,Y)\leftarrow cons(X,Z), pred1(Z,Y)$\\
$pred1(X,Y) \leftarrow Length(X,Z), succ(Z,Y)$\\
\\
We used these templates:
\begin{align*}
F(X,X) &\leftarrow F(X) & (8)\\
F(X,Y) &\leftarrow F(X,Z), F(Z,Y) & (3)\\
F(X,Y) &\leftarrow F(X,Z), F(Z,Y) & (3)
\end{align*}

\subsubsection{Son}
We aim to learn $sonOf(X,Y)$ relation from family-related facts involving $fatherOf$, $brotherOf$ and $sisterOf$. We required at least one extra intentional predicate that learns the relation $is-male$. One possible solution would be:\\
\\
$sonOf(X,Y)\leftarrow father(Y,X), isMale(X)$\\
$isMale(X)\leftarrow brother(X,Z)$\\
$isMale(X) \leftarrow father(X,Z)$\\
\\
We used these templates:
\begin{align*}
F(X,Y) &\leftarrow F(Y,X), F(X) & (11)\\
F(X) &\leftarrow F(X,Z) & (12)\\
F(X) &\leftarrow F(X,Z) & (12)
\end{align*}

\subsubsection{Grandparent}
The goal of this task is to infer the grandparent relation from observed mother-of and father-of facts. Our templates were:
\begin{align*}
F(X,Y) &\leftarrow F(X,Z), F(Z,Y) & (3)\\
F(X,Y) &\leftarrow F(X,Y) & (5)\\
F(X,Y) &\leftarrow F(X,Y) & (5)
\end{align*}

\subsubsection{Relatedness}
$related(X,Y)$ is true if there is an undirected path between $X$ and $Y$. Background knowledge contains family related facts as in the tasks Son and Grandparent.  We used these templates:
\begin{align*}
F(X,Y) &\leftarrow F(X,Y) & (5)\\
F(X,Y) &\leftarrow F(X,Y) & (5)\\
F(X,Y) &\leftarrow F(X,Z), F(Z,Y) &(3)\\
F(X,Y) &\leftarrow F(Y,X) &(9)
\end{align*}

\subsubsection{Father}
In this task we aim to learn the $Father$ relation in challenging set up (incomplete background knowledge and irrelevant facts). For details please refer to \cite{DBLP:journals/corr/abs-1711-04574}. We used these template:
\begin{align*}
F(X,Y) &\leftarrow F(X,Z), F(Z,Y) & (3)
\end{align*}

\subsubsection{Undirected Edge}
In this task the background knowledge is composed of several $edge(X,Y)$ facts. The goal is to learn $undirectedEdge(X,Y)$ which is true if there is an edge between nodes $X$ and $Y$ regardless of the direction. We used the templates:
\begin{align*}
F(X,Y) &\leftarrow F(X,Y) & (5)\\
F(X,Y) &\leftarrow F(Y,X) &(9)
\end{align*}

\subsubsection{Adjacent to Red}
In this example we extend the background knowledge of the example above with color facts: $green(C)$, $red(C)$, as well as $colour(X,C)$ which is true if node $X$ is of colour $C$. We included one auxiliary predicate that learns the relation $isRed(X)$. One possible solution would be:\\
\\
$adjToRed(X)\leftarrow edge(X,Y), isRed(Y)$\\
$isRed(X)\leftarrow colour(X,Y), red(Y)$\\
\\
We used these templates:
\begin{align*}
F(X) &\leftarrow F(X,Z), F(Z) &(13)\\
F(X) &\leftarrow F(X,Z), F(Z) &(13)
\end{align*}

\subsubsection{Two Children}
Here we aimed to learn the \textit{has-at-least-two-children(X)} predicate, which is true if there are at least two facts of the form $edge(X,Z)$. The background knowledge includes $edge$ and $neq$ (not equals) relations. We included one auxiliary predicate $pred1$. One possible solution would be:\\
\\
$twoChildren(X)\leftarrow edge(X,Y), pred1(X,Y)$\\
$pred1(X,Y)\leftarrow edge(X,Z), neq(Z,Y)$\\
\\
We used these templates:
\begin{align*}
F(X,Y) &\leftarrow F(X,Z), F(Z,Y) &(3)\\
F(X,X) &\leftarrow F(X,Z), F(X,Z) & (15)
\end{align*}

\subsubsection{Graph Colouring}
The task is to learn the \textit{adj-to-same(X,Y)} which is true if nodes $X,Y$ are of the same colour and there is an edge between them. The background knowledge is similar as in the task \textit{Adjacent to Red}. We included an auxiliary predicate that should learn the relation \textit{same-colour(X,Y)}.One possible solution would be:\\
\\
$adjToSame(X,Y)\leftarrow edge(X,Y), sameColour(X,Y)$\\
$sameColour(X,Y)\leftarrow colour(X,Z), colour(Y,Z)$\\
\\
\begin{align*}
F(X,Y) &\leftarrow F(X,Z), F(Y,Z) &(10)\\
F(X,X) &\leftarrow F(X,Z), F(X,Z) & (15)
\end{align*}

\subsubsection{Connectedness}
In this task we want to learn \textit{connected(X,Y)} which is true if there is a sequence of edges connecting nodes $X$ and $Y$. We used the templates:
\begin{align*}
F(X,Y) &\leftarrow F(X,Y) & (5)\\
F(X,Y) &\leftarrow F(X,Z), F(Z,Y) &(3)
\end{align*}

\subsubsection{Graph Cyclicity}
In this task the algorithm should learn the cocept of cyclicity. This is true of a node when there is a path departing from it and arriving to itself. The templates used were:
\begin{align*}
F(X) &\leftarrow F(X,X) & (4)\\
F(X,Y) &\leftarrow F(X,Y) & (5)\\
F(X,Y) &\leftarrow F(X,Z), F(Z,Y) & (3) 
\end{align*}


\subsection{Countries}
We mimicked some of the templates found in the appendix E of \cite{DBLP:journals/corr/RocktaschelR17}. In order to closely follow their approach, only for this task we enforced some predicates to be the same, so that \#1 represents the same predicate across the rule. In detail for each task:

\subsubsection{Countries S1}
\begin{align*}
\#1(X,Y) &\leftarrow \#1(Y,X) \\
\#1(X,Y) &\leftarrow \#2(X,Z), \#2(Z,Y)
\end{align*}
\textbf{Countries S2}
\begin{align*}
\#1(X,Y) &\leftarrow \#1(Y,X) \\
\#1(X,Y) &\leftarrow \#2(X,Z), \#3(Z,Y)
\end{align*}
\textbf{Countries S3}
\begin{align*}
\#1(X,Y) &\leftarrow \#1(Y,X) \\
\#1(X,Y) &\leftarrow \#2(X,Z), \#3(Z,W), \#4(W,Y)
\end{align*}

\subsection{Taxonomy and Kinship}

\subsubsection{Taxonomy}
\begin{figure}[!ht]
  \centering
  \includegraphics[scale=.3]{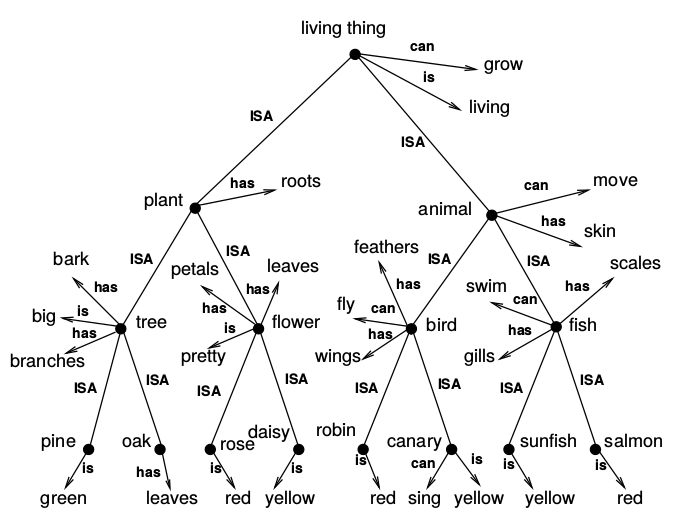}
  \caption{Bigger Animal Taxonomy used for the tasks. Contains 4 predicates, 36 constants and 145 facts}
  	\label{fig:taxonomy-big}
\end{figure}

This task was described in section \ref{taxonomy}, the dataset can be seen in figure \ref{fig:taxonomy-big}. We used the following templates to perform this task:

\begin{align*}
F(X,Y) &\leftarrow F(X,Z), F(Z,Y) & (3) \\
F(X,Y) &\leftarrow F(X,Z), F(Z,Y) & (3) \\
F(X,Y) &\leftarrow F(X,Z), F(Z,Y) & (3) \\
F(X,Y) &\leftarrow F(X,Z), F(Z,Y) & (3) 
\end{align*}

\pagebreak

\subsubsection{Kinship}
\begin{figure}[!ht]
  \centering
  \includegraphics[scale=.4]{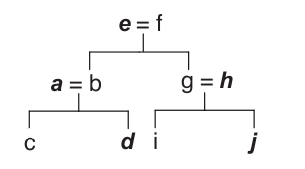}
  \caption{Inferred family tree. Females shown in bold italics and males in ordinary font.}
  	\label{fig:family}
\end{figure}

This task was described in section \ref{kinshipSec}. The theory can be seen in figure \ref{fig:family}

\end{document}